%% file: main.tex
\documentclass[runningheads]{llncs}

\usepackage{eccv}

\usepackage{eccvabbrv}

\usepackage{graphicx}
\usepackage{booktabs}
\usepackage{amsmath}
\usepackage{amssymb}
\input{latex/macros}
\usepackage[accsupp]{axessibility}  

\usepackage[pagebackref,breaklinks,colorlinks,citecolor=eccvblue]{hyperref}

\usepackage{orcidlink}
\usepackage[capitalize]{cleveref}
\crefname{section}{Sec.}{Secs.}
\Crefname{section}{Section}{Sections}
\Crefname{table}{Table}{Tables}
\crefname{table}{Tab.}{Tabs.}

\newcommand{\methodname}{Scene-LLM}

\begin{document}

\title{\methodname: Extending Language Model for \\ 3D Visual Understanding and Reasoning}

\titlerunning{\methodname}

\author{Rao Fu\inst{1}\thanks{Work done as an intern at Meta AI.} \and
Jingyu Liu\inst{2}\thanks{Work done as an AI resident at Meta AI.} \and
Xilun Chen\inst{3} \and
Yixin Nie\inst{3} \and
Wenhan Xiong\inst{3}
}

\authorrunning{R. Fu et al.}

\institute{Brown University \and
ETH Zurich \and
Meta AI}

\maketitle

\input{latex/sections/0_abstract}
\input{latex/figtab_text/fig_teaser}
\vspace{-1cm}
\input{latex/sections/1_introduction}

\input{latex/sections/2_background}
\input{latex/figtab_text/fig_1_pipeline}
\input{latex/sections/3_method}

\input{latex/figtab_text/fig_2_results_qa}
\input{latex/figtab_text/tab_0_results_qa}
\input{latex/figtab_text/tab_1_results_sqa}
\input{latex/figtab_text/tab_1_results_plan}
\input{latex/sections/4_experiment}
\input{latex/sections/6_conclusion}

%
%
\bibliographystyle{splncs04}
\bibliography{main}

\clearpage
\newpage
\input{latex/supp/all}

\end{document}

%% file: latex/macros.tex
\usepackage{multirow}
\usepackage{xcolor}
\usepackage{amsmath,amssymb,stmaryrd}   
\usepackage{siunitx}
\usepackage{array}
\usepackage{subcaption}

\usepackage{wrapfig}
\usepackage{ifthen}

\usepackage{xspace}

\newcommand{\filluptopage}[1]{%
  \clearpage
  \loop\ifnum\value{page}<#1\relax
    \null\clearpage
  \repeat
  \loop\ifnum\value{page}=#1\relax
    \null\clearpage
  \repeat
}

\makeatletter
\def\blfootnote{\xdef\@thefnmark{}\@footnotetext}
\makeatother

%% file: latex/sections/0_abstract.tex
\begin{abstract}
\vspace{-0.5cm}
This paper introduces \methodname{}, a 3D-visual-language model that enhances embodied agents' abilities in interactive 3D indoor environments by integrating the reasoning strengths of Large Language Models (LLMs). \methodname{} adopts a hybrid 3D visual feature representation, that incorporates dense spatial information and supports scene state updates. The model employs a projection layer to efficiently project these features in the pre-trained textual embedding space, enabling effective interpretation of 3D visual information. Unique to our approach is the integration of both scene-level and egocentric 3D information. This combination is pivotal for interactive planning, where scene-level data supports global planning and egocentric data is important for localization. Notably, we use egocentric 3D frame features for feature alignment, an efficient technique that incorporates the model with fine-grained concepts. Our experiments with \methodname{} demonstrate its strong capabilities in dense captioning, question answering, and interactive planning. We believe \methodname{} advances the field of 3D visual understanding and reasoning, offering new possibilities for sophisticated agent interactions in indoor settings.  
\end{abstract}

%% file: latex/figtab_text/fig_teaser.tex
\begin{figure*}[!t]
\begin{center}
\includegraphics[width=0.98\linewidth]{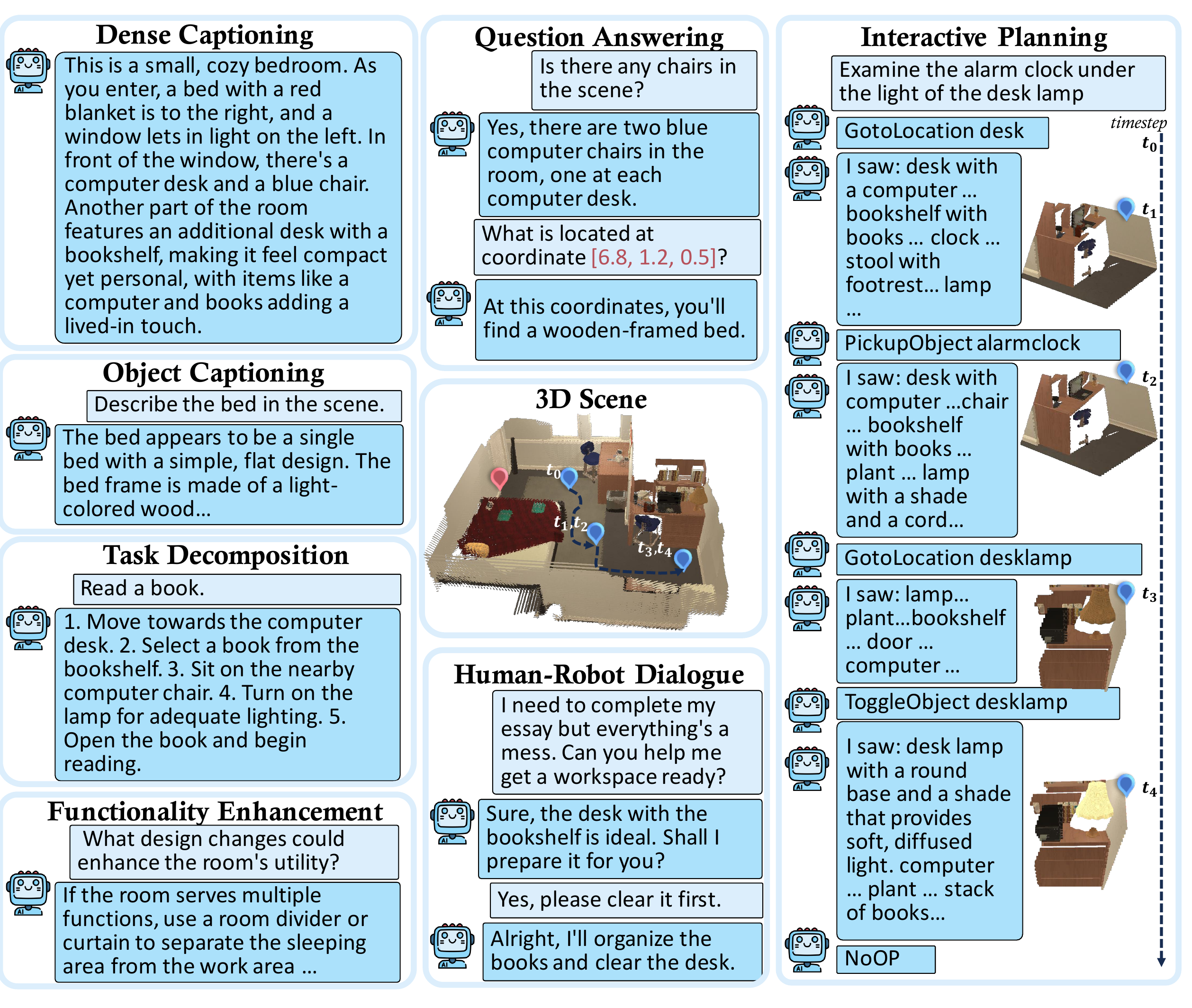}
\end{center}
\vspace{-0.5cm}
\caption{\textbf{An interactive 3D indoor scene example from an iThor\cite{Kolve2017AI2THORAn} setup.} \methodname{} is a 3D-visual-language model that can process both ego-centric and scene-level 3D visual data. We showcase some applications, including describing scene details (dense captioning), identifying and describing objects (object captioning), breaking down complex tasks into simpler steps (task decomposition), enhancing safety/utility/comfort (functionality enhancement), question-answering, generating human-robot dialogues, and interactive planning. 
}
\vspace{-0.5cm}
\label{fig:teaser}
\end{figure*}

%% file: latex/sections/1_introduction.tex
\section{Introduction}

3D visual information is important for a wide range of  tasks in indoor scenes, encompassing both egocentric tasks such as object interaction\cite{jiang2022vima} and embodied question answering\cite{ma2022sqa3d}, and broader scene-level tasks like navigation\cite{anderson2018vision} and long-horizon planning\cite{shridhar2020alfred}. While existing visual-language models (VLMs)\cite{alayrac2022flamingo, li2023blip, driess2023palm} have made strides in 2D visual-language understanding, their limited grasp of persistent 3D spatial information often renders them less effective compared to those using 3D representations for indoor scene tasks\cite{ma2022sqa3d}. 
Recent efforts\cite{ma2022sqa3d,hong20233dllm,zhu20233d} to bridge 3D visual information with textual and other modalities\cite{guo2023point} show potential in 3D visual understanding and reasoning. However, they primarily handle static 3D scenes, which is less adaptable for interactive planning involving scene changes.
In this paper, we introduce \methodname{}, a 3D-visual-language model (3D-VLM) with large language models (LLMs) as it's backbone, to address a spectrum of 3D visual understanding and reasoning tasks within interactive scenes, ranging from dense captioning to interactive planning, as illustrated in Figure~\ref{fig:teaser}.

For 3D scene tasks, we recognize the importance of both egocentric and comprehensive scene-level information. Egocentric information is crucial for immediate updates during object interactions and for localizing the agent within the scene. Conversely, scene-level information provides temporal persistent and multi-view consistent details of the entire 3D scene. This is essential for map-reliant tasks such as navigation and long-horizontal planning. Current models\cite{driess2023palm,ma2022sqa3d,hong20233dllm} typically focus on one of these aspects, hindering their effectiveness in tasks like interactive indoor planning that require both. To overcome this, we propose integrating both types of 3D visual information in \methodname{}. For egocentric information representation, previous works have employed images\cite{driess2023palm}, or agent coordinates\cite{ma2022sqa3d}  necessitating multiple modality encoders. In contrast, we employ 3D point sets as a uniform feature representation, capable of encompassing 2D egocentric information by projecting depth images to 3D frames and representing scene-level information by aggregating 3D frames or sampling points on the 3D scene's surface. Instilling these data allows for the interpretation of both information sources during interactive planning, as depicted in Figure~\ref{fig:teaser}.

Another key issue is aligning the dense 3D visual information with the textual embedding space of a pre-trained LLM. 
While previous studies have explored using patchy visual features\cite{liu2023visual} and query tokens\cite{li2023blip} for downsampling visual features, 3D point sets pose a unique problem due to their continuous coordinate system and the need for a representation that can adapt to changes in scene state.
To address this, we employed a hybrid representation that spatially downsamples the 3D point set using resolution-fixed voxel grids. This method not only retains dense spatial information but also facilitates interactive updates. This varied length prepresentation has proved effective across various tasks, indicating the model is able to understand spatial coordinates effectively.

In aligning textual and visual features, we followed a two-stage strategy similar to previous studies\cite{liu2023visual}. We used an image caption model\cite{chen2023minigpt} and a LLM\cite{touvron2023llama} to generate conceptual and instructional following annotations. Initially, we trained the projection layer using the conceptual annotations while keeping the LLM frozen, and then fine-tuned it with instructional following annotations alongside the LLM. During the training of the projection layer, we noted that relying solely on generated scene caption data could cause the model to miss small object details. Rather than incorporating additional object datasets as done in previous work\cite{hong20233dllm}, we directly used 3D frame data in two coordinate systems, camera and world, to fine-tune the projection layer. This data includes numerous concepts relating to small-scale objects, and the frame features themselves contain more fine-grained details. Empirically, we found that training with 3D frame data led to faster convergence than using scene captions or instructional following data.

Empirical evaluations demonstrate that \methodname{} excels in a wide range of 3D scene reasoning tasks, achieving state-of-the-art results on the ScanQA\cite{azuma2022scanqa} and SQA3D benchmarks\cite{ma2022sqa3d} without additional fine-tuning. Thanks to its capability of understanding egocentric and scene-level information, \methodname{} effectively handles scene changes, enabling interactive planning in dynamic environments. When fine-tuned for specific tasks, \methodname{} outperforms other LLM-based models on the Alfred benchmark\cite{shridhar2020alfred}. In summary, our primary contributions are: 
\begin{itemize}
    \item We introduce \methodname{}, a 3D-VLM that connecting 3D visual information with LLM and sets new state-of-the-art on 3D-VQA and interactive planning benchmarks;
    \item We propose an effective 3D visual representation that captures fine-grained 3D information and support state change by design. This representation can be easily incorporated into LLMs with a light-weighted projector;
    \item We create a large-scale dataset which is useful for 3D and text feature alignment, which includes $190k$ 3D-visual-language pairs from an egocentric viewpoint and about $500k$ pairs of scene-level data.
\end{itemize}


%% file: latex/sections/2_background.tex
\section{Related Works}
\label{sec:background}
\subsection{3D Scene and Language Benchmarks.}
Various benchmarks have been developed to explore the intersection of 3D scenes and language. These include tasks like text-guided scene synthesis\cite{tan2019text2scene, hwang2023text2scene}, caption and description generation\cite{achlioptas2020referit3d, chen2021scan2cap}, object grounding and reference\cite{chen2020scanrefer, achlioptas2020referit3d, feng2021free, huang2021text, zhang2023multi3drefer}, 3D visual question answering (3D-VQA)\cite{ye20223dqa,ma2022sqa3d,azuma2022scanqa,hong20233dclr,hong20233dllm, zhu20233d}, interactive indoor planning\cite{puig2018virtualhome, shridhar2020alfworld, shridhar2020alfred}, and Visual-Language Navigation (VLN)\cite{anderson2018vision, krantz2020beyond, huang2023visual}. Our work is focused on integrating 3D scene understanding and reasoning with linguistic modalities, a key element for developing indoor scene agents and harnessing the reasoning capabilities of pre-trained LLMs. We concentrate on benchmarks involving textual outputs such as captioning, 3D-VQA, and high-level indoor planning tasks. \methodname{} demonstrates an understanding of spatial coordinates through bootstrap training with spatial data. However, for this stage, we defer coordinate-level spatial understanding and planning to specialized, existing models\cite{blukis2021a}.
\subsection{Interactive Indoor Planning.}
Interactive indoor planning involves agents navigating and performing tasks within interactive indoor environments, based on language instructions. While multi-modality models mapping language to actions have shown promise \cite{misra2017mapping, suglia2021embodied}, they often lag behind hierarchical planning approaches \cite{zhang2021hierarchical, blukis2021a, min2022film}, which segment tasks into high-level strategy and low-level execution.
Recent advancements leverage LLMs for planning, employing them for concept generation \cite{shah2023lm}, as direct planners \cite{ahn2022can, lu2022neuro, huang2022language, zheng2022jarvis, song2022llm, chalvatzaki2023learning, xie2023translating, liu2023picture, liu2023reflect, singh2023progprompt, rana2023sayplan}, or through fine-tuning with embodied data \cite{xiang2023language}. However, these models typically require supplementary visual understanding modules. 
Some approaches \cite{driess2023palm, jiang2022vima, reed2022generalist} treat visual features as textual tokens for LLMs but are limited to current state information, lacking a comprehensive environmental perspective. Other methods \cite{huang2023visual, huang2023voxposer} utilize dynamic visual-language semantic maps (VLMs) in 2D or 3D to enhance LLM planning capabilities.
\methodname{} employs 3D scene-level visual features as VLMs, along with 3D frame features for updated visual inputs, which are directly fed into the LLM. This process instills dense 3D visual information in the LLM and streamlines the planning process.
\subsection {3D Scene Representation.}
3D scene representation is broadly classified into three categories. The first is object-centric representation, including bounding boxes\cite{qi2019deep}, object category names, and object features\cite{sajjadi2022object}. These representations are intuitive and computationally efficient, but require additional design to track objects in interactive environments and often overlook dense information. Another category involves dynamic graphs\cite{li2022pre, agia2022taskography, kurenkov2023modeling}. These are memory-efficient and adept at tracking changes in interactive scenes. The challenge lies in their complex topology, which makes intuitive integration with LLMs difficult. The third category is dense representation, which includes spatial maps\cite{blukis2021a, huang2023visual, chen2023open}, voxel grids\cite{wang2017cnn, riegler2017octnet, liu2019point}, point-wise\cite{jatavallabhula2023conceptfusion}, and field-like formats\cite{mescheder2019occupancy, mildenhall2021nerf, kerr2023lerf, hong20233dclr}. Some approaches\cite{qi2017pointnet, jiang2020pointgroup, zhao2021point, wang2023octformer} enhance effectiveness by grouping or tokenizing scenes, although this often depends heavily on task-specific supervision. In our work, \methodname{} employs a point-wise representation to preserve dense information effectively. We incorporate voxelized spatial downsampling\cite{liu2019point} to manage computational demands and facilitate visual feature updates.
\subsection{Visual-Language Models (VLMs).}
The development of general 2D VLMs has gained significant momentum\cite{tsimpoukelli2021multimodal, alayrac2022flamingo, li2023blip, driess2023palm}. These models integrate pretrained vision models with pretrained LLMs using extensive datasets. Some studies\cite{yang2023mm, liu2023visual, zhu2023minigpt} create text-visual data pairs by prompting other LLMs or VLMs, broadening language applications to instruction-specific formats. Recent works\cite{wang2023visionllm, peng2023kosmos, chen2023minigpt, you2023ferret} also demonstrate VLMs' capability in spatial understanding, either by designing spatial tokens or augmenting text with spatial descriptions. There is also a growing interest developing 3D-VLM\cite{hong20233dllm, zhu20233d, guo2023point}, with approaches like point resampling\cite{hong20233dllm}, LLM adaptors\cite{guo2023point}, and multi-modality transformers\cite{zhu20233d} to bridge 3D and textual modality. Unlike these methods, our work, \methodname{}, diverges in its focus on interactive scene tasks. We employ distinct feature representations, model structures, alignment data, and tuning strategies tailored to interactive 3D environments.

%% file: latex/figtab_text/fig_1_pipeline.tex
\begin{figure*}[!t]
\begin{center}
\includegraphics[width=0.99\linewidth]{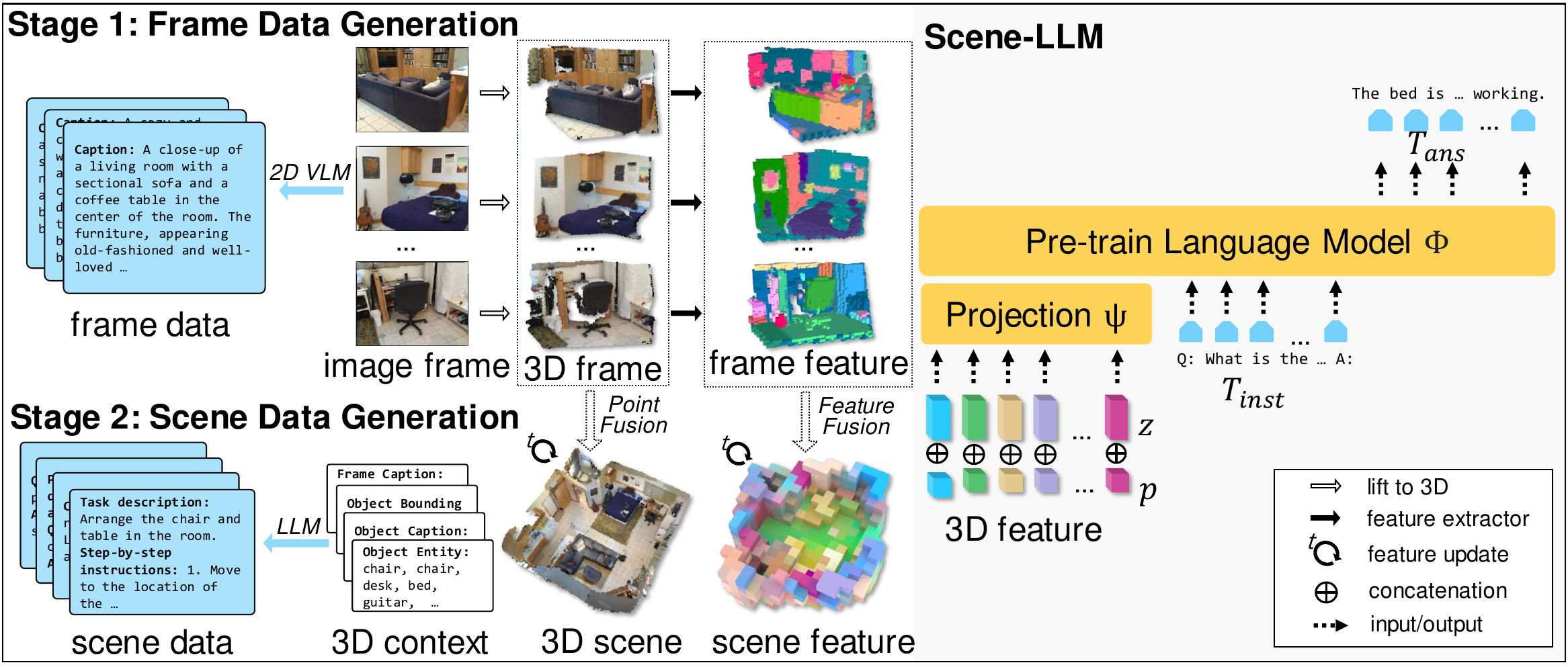}
\end{center}
\vspace{-0.5cm}
\caption{\textbf{Overview of the data generation process and \methodname{}'s architecture.} The data generation comprises two stages: a 3D frame-language generation stage, which uses image frames and a 2D VLM to generate frame descriptions, and a 3D scene-language data generation stage, which uses 3D context data and LLM to generate scene instruction-following data. Scene-LLM bridge takes tokenized 3D semantic features concatenated with 3D coordinates as input. It bridges 3D visual features with textual feature space with a simple projection layer. 
}
\vspace{-0.5cm}
\label{fig:pipeline}
\end{figure*}

%% file: latex/sections/3_method.tex
\section{3D-Visual-Language Data Generation}
To align 3D visual information with LLMs, we first prepare paired 3D-visual-language data. Our dataset comprises about $9,000$ indoor scenes from three sources: real indoor scans\cite{Dai_2017_CVPR}, single rooms from the Habitat-Matterport 3D dataset (hm3d)\cite{ramakrishnan2021habitat}, and expert-designed scenes from iThor\cite{Kolve2017AI2THORAn}. Recognizing the importance of egocentric and scene-level information in interactive planning tasks, we generated two types of data: 3D frames paired with language annotations for egocentric understanding, and 3D scenes paired with language annotations for scene-level understanding.

\subsection{Frame Data Generation.}
To generate 3D frame data, we render scenes from multiple camera perspectives, capturing RGB frames and ground truth depths. Each frame's pixels are projected into spatial coordinates, forming a 3D point set – referred to as a 3D frame. We have amassed approximately $100k$ 3D frames from our dataset. For paired language annotations, we utilize MiniGPT-V2\cite{chen2023minigpt}, capable of generating captions and object descriptions from images by using caption and grounded caption identifiers. We have collected around $190k$ 3D-frame-language pairs, essential for instilling fine-grained concept understanding and egocentric awareness in \methodname{}. Details on the data generation process are provided in the supplementary material.

\subsection{Scene Data Generation.}
For \methodname{} to comprehend scene-level information, we produce paired 3D-scene-language data. This data includes a broader perspective of the scene and diverse instructional types to equip \methodname{} with various capabilities. 3D scene data is reconstructed by aggregating 3D frames based on their camera poses, ensuring visual feature consistency with the frame data. Language annotations for scenes are generated using Llama-2-chat-70B\cite{touvron2023llama}, prompted with a mix of context data including generated frame captions, frame object descriptions, annotated object lists, and annotated bounding boxes. These prompts lead to diverse instruction-following data types like dense caption, object caption, task decomposition, functionality enhancement, question-answering, and human-robot dialogues. Acknowledging the potential noise in generated data, we apply a self-checking method\cite{zhu2023minigpt} to refine data quality. Approximately $500k$ instruction-following data pairs are generated. The prompts and instructional data distributions are provided in the supplementary material. Importantly, as we use bounding box information as scene context, some captions and QA data include spatial coordinates. We intentionally retain these data to partially imbue \methodname{} with spatial coordinate understanding.

\section{\methodname{}}
\methodname{} is a 3D Visual Language Model (VLM) with a simple yet effective architecture designed to comprehend both egocentric and scene-level 3D visual information, enabling it to successfully perform interactive planning tasks. This section outlines the 3D visual feature extraction process, our model's architecture, the alignment of 3D visual information with our dataset, and the use of \methodname{} for inference.

\subsection{3D Visual Feature}
\label{sec:feature}

In aligning visual and language modalities, dual-modality semantic features have proven effective. Following prior research \cite{ramesh2022hierarchical, liu2022instruction, li2023blip, zhu2023minigpt}, we employ visual-language semantic features \cite{radford2021learning} to represent 3D visual semantics. This involves first extracting pixel-wise CLIP features from each image and then aggregating these into a 3D point set, as inspired by ConceptFusion \cite{jatavallabhula2023conceptfusion}. This process transforms pixel-wise features into point-wise features, capturing the semantic nuances of 3D visual data. Each point in our 3D frame data corresponds to a pixel-wise feature, and each point in the 3D scene data corresponds to a point-wise feature within this aggregated set, ensuring consistency between the frame and scene features.

To tokenize these 3D visual features for compatibility with LLM input, we adopt a hybrid point-voxel representation \cite{liu2019point}, balancing the need for dense 3D visual information, support for interactive updates, and manageable token lengths for the LLM. For each set of 3D visual data(from either 3D frames or scenes), we first divide the space into a fixed-resolution voxel grid with dimensions $X\times Y\times Z$, where $X, Y, Z$ represent the voxel counts along respective axes. The number of voxels varies across different scenes due to this fixed resolution. Secondly, for each voxel, we cluster any contained points using a K-Nearest Neighbors (KNN) approach. Each point's features include semantic attributes and spatial coordinates. We then compute the average of the largest cluster's features within each voxel, yielding a voxel grid of features  ${F}^{Vox}\in \mathbb{R}^{X\times Y\times Z \times(D+3)}$, where $D$ is the dimension of the semantic feature $z$ and $3$ is the dimension of the spatial coordinates $p$. Lastly, A visibility map $V \in \{0, 1\}^{X\times Y\times Z}$ is computed, indicating the presence(1) or absence(0) of points in each voxel. Only features from visible voxels are used as visual tokens, which means \methodname{} takes a variable number of tokens for different scenes. 

This hybrid representation retains dense spatial information by uniformly downsampling the point set while facilitating feature updates akin to map representations \cite{blukis2021a,huang2023visual}. To update scene features $\textbf{F}^{Vox}_t$ at state $t$ to state $t+1$, we first render a 3D frame from the current camera view. This frame's semantic features $f$ are projected to a 3D point-wise feature map $\hat{\textbf{F}}$ and voxelized into $\hat{\textbf{F}}^{Vox}$ with a corresponding visibility map $\hat{V}$. The scene semantic feature is then updated using:
\begin{equation}
    \label{equation:update}
    \textbf{F}^{Vox}_{t+1} = \textbf{F}^{Vox}_{t} \times(1 - \hat{V}) + \hat{\textbf{F}}^{Vox} \times \hat{V},
\end{equation}
ensuring that the semantic representation of the 3D scene remains synchronized with any scene state changes.

\subsection{3D-Visual-Language Alignment}
\noindent\textbf{Model Architecture}
\label{sec:align}
Building on the concept of visual instruction tuning \cite{liu2022instruction}, we have developed a unified 3D visual-language model that effectively aligns 3D visual information with an LLM. Our framework utilizes Llama-2-7b \cite{touvron2023llama} as the foundational LLM backbone, denoted by $\Phi$. 
To bridge 3D visual tokens($\textit{F}$) with the LLM's tokenized space, we introduce a linear projection layer($\psi$), transforming $\textit{F}$ into aligned 3D visual tokens($\textit{T}_{3D} = \psi(\textit{Z})$), where $\textit{T}_{3D} \in \mathbb{R}^{H}$ aligns with the LLM's feature dimension ($H$).

\noindent\textbf{Stage 1: Pretraining for Feature Alignment}
In the initial pretraining phase, we exclusively use 3D frame-language data. We utilize 3D frame data under two coordinate systems—camera and world coordinates—to ensure that \methodname{} comprehends both egocentric and scene-centric perspectives. This dual coordinate system aids in grasping fine-grained visual details and more textual concepts, respectively. Notably, training with frame data under these two coordinate systems proved as effective as using scene caption data, with faster convergence observed. The paired 3D-frame-language data are then processed with special tokens \texttt{[3D]} and \texttt{[/3D]} demarcating the modalities. At this stage, only the projection layer is trained, allowing efficient alignment of 3D visual features with textual features, while keeping the LLM parameters ($\Phi$) unchanged.

\noindent\textbf{Stage 2: Finetuning}
In the finetuning phase, \methodname{} is optimized to respond accurately to user instructions. We incorporate both 3D frame-language and 3D scene-language data, using the identifier token \texttt{"I saw"} to preface all 3D frame-language data. Textual descriptions are bifurcated into instructions($\textit{T}_{inst}$) and their corresponding responses($\textit{T}_{ans}$). Leveraging the transformed 3D visual token($\textit{T}_{3D}$) and the instruction token($\textit{T}_{inst}$), our goal is to fine-tune the LLM($\Phi$) to autoregressively generate $\textit{T}_{ans}$:
\begin{equation}
    p(\textit{T}_{ans}; \textit{T}_{3D}, \textit{T}_{inst}) = \prod_{i=1}^{L}p_{\theta}(t_i | \textit{T}_{3D}, \textit{T}_{inst}, \textit{T}_{ans, <i}),
\end{equation}
where $L$ represents the total token length, $t_i$ is the $i^{th}$ token, and $\theta$ denotes the tunable parameters. Here, we jointly fine-tune both the projection layer and the LLM, indicated by $\theta = \{\psi, \Phi\}$.

\subsection{Inference}
\label{sec:inferece}
During inference, for non-interactive tasks, \methodname{} processes the 3D visual data and user instructions to autoregressively generate responses. For interactive tasks, we adopt a two-step inference approach, including an egocentric and a scene-level update step. At the egocentric step, 3D frame data and a egocentric instruction are first input to \methodname{} to describe the current state. The egocentric instruction is an ``I saw'' identifier.
Then, the 3D scene feature is updated as per Equation \ref{equation:update}. The updated scene feature, along with the state description and user instructions, are fed into \methodname{} to yield the corresponding response. The instruction is composed of the \textbf{current frame description}, task description, \textbf{completed steps}, and the ``Next-step:'' identifier, with \textbf{bold} parts being updated per timestep. These updates are vital for observation collection, agent grounding, and replanning.

%% file: latex/figtab_text/fig_2_results_qa.tex
\begin{figure*}[!t]
\includegraphics[width=0.99\linewidth]{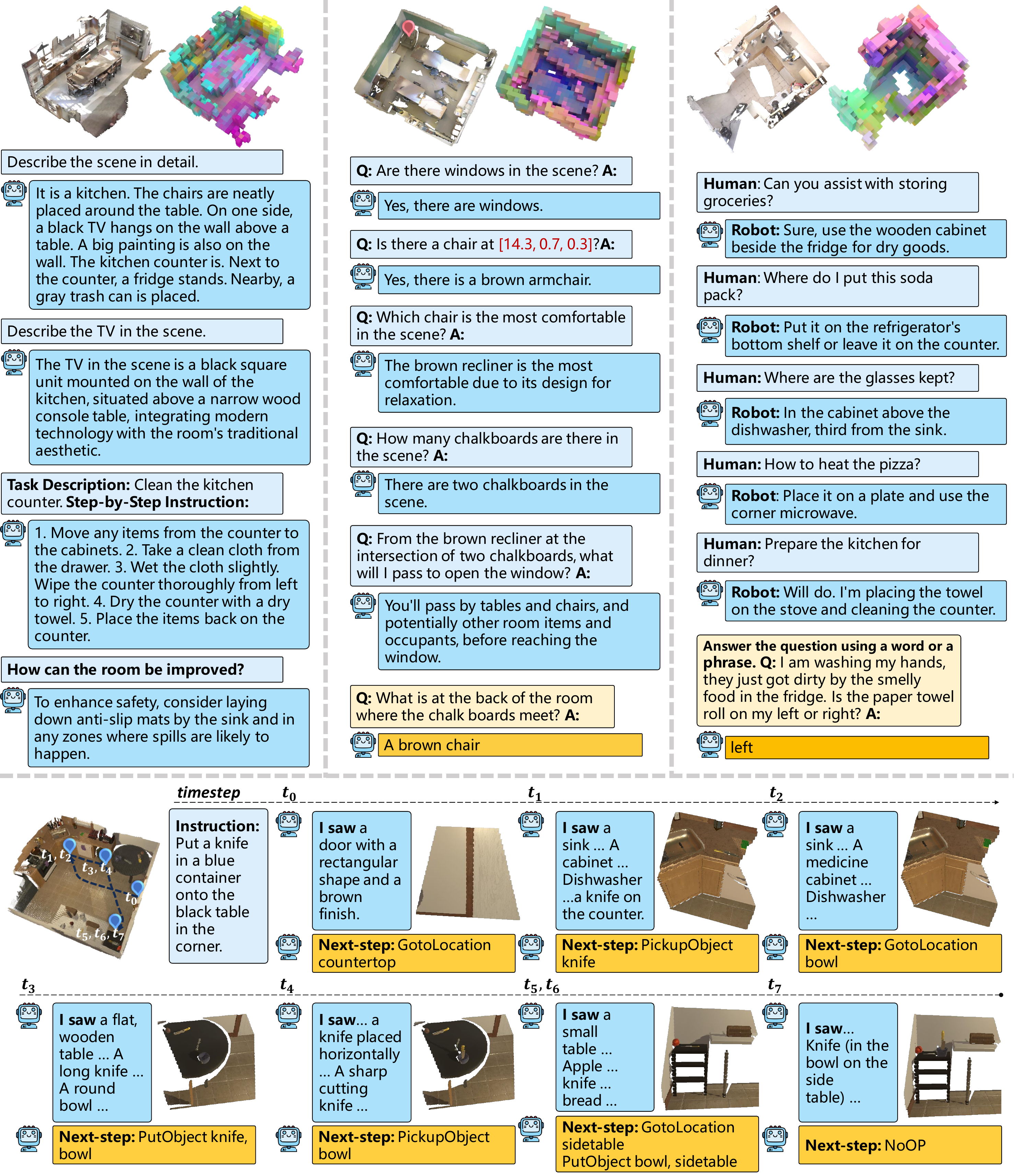}
\caption{Examples using \methodname{} for non-interactive and interactive tasks. On the right of the static scenes are visualization of input scene features. Bold texts are task-specific identifiers. Yellow boxes are the results from the ScanQA/SQA3D/Alfred benchmark. Responses for the top $3$ non-interactive scenes are generated without task-specific finetuning, and those for the bottom interactive scene are generated with finetuning. Examples show that \methodname{} can fulfill multiple tasks and is equipped with 3D visual understanding and reasoning abilities. }
\label{fig:result_qa}
\end{figure*}

%% file: latex/figtab_text/tab_0_results_qa.tex
\begin{table*}[t]
\centering
\small
\tabcolsep 2pt
\caption{
Performance on ScanQA benchmark validation set. Metric reported include Exact Match, BLEU, ROUGE-L, METEOR, and CIDEr. The `*' symbol indicates task-specific fine-tuning. The bold text highlights the best results. \methodname{} performs the best among most metrics. \methodname{} performs the best among most of the metrics.
}
\label{tab:scanqa}
\vspace{-0.2cm}
\begin{tabular}{lcccccccc}
\toprule
Method & \multicolumn{1}{c}{{EM@1}} & \multicolumn{1}{c}{{B-1}} & \multicolumn{1}{c}{{B-2}}& \multicolumn{1}{c}{{B-3}}& \multicolumn{1}{c}{{B-4}} & \multicolumn{1}{c}{{METEOR}} & \multicolumn{1}{c}{{ROUGE-L}} & \multicolumn{1}{c}{{CIDEr}}\\
\midrule
VoteNet+MCAN\cite{yu2019deep} & 17.3  & 28.0 & 16.7 & 10.8 & 6.2 & 11.4 & 29.8 & 54.7 \\
ScanRefer+MCAN\cite{yu2019deep} & 18.6 & 26.9 &16.6 &11.6 &7.9 & 11.5 & 30 & 55.4 \\
ScanQA \cite{azuma2022scanqa} & 21.0 & 30.2 & 20.4 & 15.1 & 10.1 &  13.1 & 33.3 & 64.9 \\
3D-LLM \cite{hong20233dllm} & 20.5 & 39.3 & 25.2 & 18.4 & 12.0 & 14.5 & 35.7 & 69.4 \\
\methodname{}& {25.6} & {42.2} & {26.4} & {18.7} & 11.7 & {15.8} & {35.9} & \textbf{80.0} \\
\methodname{}*& \textbf{27.2} & \textbf{43.6} & \textbf{26.8} & \textbf{19.1} & \textbf{12.0} & \textbf{16.6} & \textbf{40.0} & \textbf{80.0} \\
\bottomrule
\end{tabular}
\vspace{-0.2cm}
\end{table*}

%% file: latex/figtab_text/tab_1_results_sqa.tex
\begin{table*}[t]
\centering
\small
\caption{
Exact Match Metric on SQA3D test set. Metric is reported under $6$ for different question types. The `*' symbol indicates task-specific fine-tuning. The bold text highlights the best results. \methodname{} performs the best among most of the metrics.
}
\label{tab:sqa3d}
\vspace{-0.2cm}
\tabcolsep 8pt
\begin{tabular}{lccccccc}
\toprule
\multirow{2}{*}{{Method}} & \multicolumn{6}{c}{{\textit{Test set}}} & \multirow{2}{*}{{Avg.}} \\ 
\cline{2-7}
& {What} & {Is} & {How} & {Can} & {Which} & {Other} & \\ 
\hline
GPT-3 & 39.7 & 46.0 & 40.5 & 45.6 & 36.1 & 38.4 & 41.0 \\ 
ClipBERT\cite{lei2021less} & 30.2 & 60.1 & 38.7 & 63.3 & 42.5 & 42.7 & 43.3 \\
SQA3D\cite{ma2022sqa3d} & 31.6 & 63.8 & \textbf{46.0} & 69.5 & 43.9 & 45.3 & 46.6 \\
3D-Vista\cite{zhu20233d} & 34.8 & 63.3 & 45.4 & 69.8 & \textbf{47.2} & 48.1 & 48.5 \\
\methodname &  {40.0} & \textbf{69.2} &  42.8 & \textbf{70.8}  & 46.6  &  \textbf{52.5} &   {53.6}\\
\methodname* &  \textbf{40.9} & {69.1} &  45.0 & \textbf{70.8}  & \textbf{47.2}  &  {52.3} &   \textbf{54.2}\\
\bottomrule
\end{tabular}
\vspace{-0.2cm}
\end{table*}

%% file: latex/figtab_text/tab_1_results_plan.tex
\begin{table}[t]
    \centering
    \small
    \tabcolsep 1.5pt
    \renewcommand\arraystretch{1.}
    \captionsetup{width=1.0\textwidth} 
    \caption{Result on Alfred dataset on test unseen/seen set and valid unseen/seen set. The metrics reported include success rate (SR), goal-conditioned success rate(GC), and high-level planning accuracy(HLP). The notation "(step)" denotes textual inputs that include step-by-step instructions. \methodname{} performs the best among methods using goal instruction only. Its HLP performs the best among all methods.
    }
    \label{tab:alfred}
    \begin{tabular}{lcccccccccc}
    \toprule
    \multirow{2}{*}{{Model}} &  \multicolumn{2}{c}{{Test Unseen}} & \multicolumn{2}{c}{{Test Seen}} &
    \multicolumn{3}{c}{{Valid Unseen}} & \multicolumn{3}{c}{{Valid Seen}}  \\
     \cmidrule(r){2-3} \cmidrule(r){4-5} \cmidrule(r){6-8} \cmidrule(r){9-11}
    &  {SR}  & {GC} & {SR} & {GC} & {SR}  & {GC} & {HLP} & {SR} & {GC} & {HLP} \\
    \midrule
    HiTUT\cite{zhang-chai-2021-hierarchical} & 11.12& 17.89& 13.63& 21.11& 10.23& 20.71& –-& 18.41& 25.27& –- \\
    HLSM\cite{blukis2021a}   & 20.27  & 27.24 & 25.11  & 35.79 & {18.28} & {31.24} & {70.17} & 29.63 & 38.74 & 77.64 \\ 
    LLM-Planner\cite{song2022llm}& 13.41 &22.89 &15.33 &24.57 &12.92 &25.35 &55.85 &13.53 &28.28 &54.33 \\
    E.T.(step)\cite{pashevich2021episodic} & 8.57 & 18.56 & {38.42} & {45.44} & 7.32 & 20.87 & -- & {46.59} & {52.92} & -- \\
    FILM(step)\cite{min2022film} & 27.80 & {38.52} & 28.83 & 39.55 &  -- & -- & 54.93 &-- & --& 60.86 \\
    LLM-Planner(step)\cite{song2022llm}& {16.42} & {23.37} & {18.20} & {26.77} & {15.36} & {29.88} & {68.31} & {16.45} & {30.11} & {71.84} \\
    \methodname &   {25.15} & {33.75}& {26.52}& {37.09}& {22.35}& {38.86}& {79.63}& {35.56}& {45.04}& {82.78} \\
     \bottomrule
    \end{tabular}
    \vspace{-0.5cm}
\end{table}

%% file: latex/sections/4_experiment.tex
\section{Experiments}
In this section, we detail our benchmark results and provide examples to illustrate \methodname{}'s capabilities in 3D visual understanding and reasoning. We conducted ablation studies to demonstrate the effectiveness of our 3D visual representation, the effectiveness of frame data, and the impact of our training strategies. Additionally, we observed that increasing spatial resolution enhances performance, suggesting further improvement direction. 

\subsection{Results and Benchmark Evaluation}
\label{sec:results}

\subsubsection{Benchmarks and Metrics.} To evaluate \methodname{} in tasks related to 3D scene reasoning and planning, we utilized three primary datasets for benchmarking. 
\textbf{ScanQA\cite{azuma2022scanqa}.} This benchmark tests a model's ability to understand 3D scenes using question-answering tasks using ScanNet dataset\cite{Dai_2017_CVPR}. Answers are typically short phrases.
\textbf{SQA3D\cite{ma2022sqa3d}.} This benchmark assesses a model's 3D scene understanding and reasoning through situated QA tasks using ScanNet scenes\cite{Dai_2017_CVPR}. The answers are generally one-word or one-phrase.
\textbf{ALFRED\cite{shridhar2020alfred}.} It measures the ability to create precise and robust plans from a high-level goal in 3D interactive environments from iTHOR\cite{Kolve2017AI2THORAn}. Results on more benchmarks are included in the Supplementary Material.

\subsubsection{Performance on 3D-VQA benchmarks.} Our evaluation of \methodname{} on 3D visual question answering (3D-VQA) benchmarks is summarized in Table \ref{tab:scanqa} for ScanQA and Table \ref{tab:sqa3d} for SQA3D, comparing it against other baseline methods. We present results both with and without task-specific finetuning for a comprehensive analysis. 

\noindent\textbf{On the ScanQA Benchmark:} \methodname{} demonstrates superior performance over other methods in most metrics, even without task-specific finetuning. This highlights its robust understanding of 3D scenes and efficient reasoning capabilities. 

\noindent\textbf{For SQA3D:} By incorporating the identifier "Answer the question using one word or one phrase," \methodname{} aligns well with the SQA3D's output format from the outset. It excels in most metrics compared to competing methods. An exception is observed in "How"-type questions, particularly counting tasks, where 3D-Vista\cite{zhu20233d} shows an edge due to its object-centric representation, which is inherently advantageous for such queries. Nevertheless, \methodname{}, with its dense representation approach, holds a strong overall performance. 

\noindent\textbf{Task-Specific Finetuning:} Our model shows strong generalizability even prior to fine-tuning. Average performance improved across benchmarks post-tuning, although minor decreases were observed in 2 question types in SQA3D\cite{ma2022sqa3d}, echoing variability seen in \cite{ma2022sqa3d,zhu20233d}. According to \cite{ma2022sqa3d} Sec.4.2, `Is' type is more guess-prone, while `What'\&`Which' types, where our model shows strength, better evaluate visual understanding. This suggests that the model's effectiveness is further amplified when tailored to specific task requirements, demonstrating its adaptability and potential for refinement.

\noindent \textbf{Performance on Interactive Planning Benchmark}
\label{sec:result_planning}
Fig.~\ref{fig:result_qa} compares the performance of \methodname{} with other methods. We finetune \methodname{} with only $2k$ steps to align its output with Alfred's high-level command format. As we only focus on the interactive high-level planning ability, which orthogonal to low-level controlling we directly adopt the low-level controller from \cite{blukis2021a} following \cite{song2022llm}. Specifically, we replace the high-level controller of HLSM with \methodname{}'s output. The inference process follows section \ref{sec:inferece}. We report success rate(SR)\cite{shridhar2020alfred}, goal-conditioned success rate(GC)\cite{shridhar2020alfred}, and high-level planning accuracy(HLP)\cite{song2022llm}. Comparing the all metrics between HLSM\cite{blukis2021a} and \methodname{}, replacing the high-level controller with \methodname{} brings performance boost among all metrics, which proves the excel of \methodname{} in high-level controlling. 
Our advantage in high-level task decomposition is evident in the HLP metric, surpassing recent high-level planning works\cite{song2022llm}, which uses extra visual feature extractor for egocentric visual feature extraction. This proves our multi-modal method achieves faithful visual feature extraction without the need for a specialized general visual feature extractor.
Notably, our methodology employs only goal instructions, yet achieves performance comparable to methods that utilize step-by-step instructions (step). This underscores the exceptional capability of \methodname{} in task decomposition.

\noindent \textbf{Qualitative Results.} 
Figure \ref{fig:result_qa} showcases examples of \methodname{}'s performance on three static scenes from the ScanNet dataset\cite{Dai_2017_CVPR} and one interactive scene from the iThor environment\cite{Kolve2017AI2THORAn}. In these examples, we also provide visualizations of the 3D visual features used by \methodname{}. These features are reduced to three dimensions via Principal Component Analysis (PCA) and color-coded based on their PCA components for clarity. The selected examples highlight the wide range of capabilities of \methodname{}: common sense understanding, concept understanding, spatial relationship understanding, coordinate understanding, counting, task decomposition, functionality enhancement, embodied reasoning, and navigation skills. The interactive scene example shows \methodname{} understanding both egocentric and scene-level information and can plan interactively in the scene without any extra visual information extractors.

\input{latex/figtab_text/tab_ablation}
\subsection{Ablation Studies and Discussions.}
We evaluate and discuss the major design choices with ScanQA and SQA3D in this section. Detailed ablations including modality comparisons, fine-tuning efficiency with frame data, and contrasts between frame and scene data, and more are available in the Supplementary Material.

\noindent \textbf{3D Visual Features.}
We ablate different 3D representations, including object aligned bounding boxes (a) and class labels detected by VoteNet\cite{qi2019deep}, object-centric point clouds features following 3D-Vista\cite{zhu20233d} (b), point clouds down-sampled (c) with farthest point sampling, point clouds queried (d) with a fine-tuned Q-Former\cite{li2023blip} and point clouds down-sampled (e) with voxel-grid. 
The findings reveal that our approach for 3D visual features either matches or surpasses other representations in performance with with reduced parameters. Interestingly, object-centric methods (a\&b) demonstrate similar effectiveness to spatial down-sampling techniques (c\&e). This could be due to most textual data in ScanQA and SQA3D being annotated based on object bounding boxes. While Q-Former is a robust down-sampling technique, it exhibits slightly lower performance compared to direct spatial down-sampling in our benchmarks, aligning with findings from \cite{liu2023visual}. 

\noindent \textbf{Training Strategy.}
In our study, we evaluated \methodname{}'s performance across different training strategies: (1) With pretraining (f) versus without pretraining (e); (2) Using frame data (g) versus scene data (f) during pretraining.
The results indicate a significant performance boost when pretraining is implemented, emphasizing the importance of concept alignment in the initial training phase. Interestingly, using frame data either matched or slightly exceeded the effectiveness of scene data. We ascribe this to the richer conceptual content found in frame data. Detailed statistics are available in the supplementary material.
\input{latex/figtab_text/tab_ablation_plan}

\noindent \textbf{Inference Strategy.} We evaluate the importance of egocentric and scene-level information update during interactive planning. As our method is orthogonal to low-level controller, the high-level planning accuracy (HLP) on Alfred serves as a direct indicator of our planning proficiency.
Table.~\ref{tab:ablation_plan} demonstrates the impact of excluding updates
from current frame description (w/o egocentric) and scene features (w/o scene). The absence of egocentric updates leads to a significant decline in HLP. This decline highlights the dual role of egocentric information: it not only provides a snapshot of the current state but also anchors the agent's understanding of that state. Similarly, excluding scene state updates results in a modest decrease in HLP, emphasizing that information updates play a valuable part in interactive planning.

\input{latex/figtab_text/fig_4_ds}
\noindent \textbf{Voxel Grid Resolution.} 
Due to the max token length of LLama2 is $4096$, we use a fixed voxel downsample resolution $0.18$ in our experiment. However, in Fig.~\ref{fig:ds} (a)(b), we show that increasing the resolution result into performance boost on QA benchmarks. This suggest that \methodname{} might be further improved by incorporating LLMs that process longer text tokens\cite{xiong2022adapting, xiong2023effective} to process dense information more effectively.

\noindent \textbf{Number of views.} During 3D visual feature extracting process, we randomly sample views from input 3D scenes. Fig.~\ref{fig:ds} (c) reports the Exact Match (EM) score on SQA3D using different number of views. Figure shows the EM score has minor increment with the number of views increases, comparing with the change of voxel resolution. This further shows the main bottleneck for scene feature extraction is the voxel resolution, which is limited by the max token length of LLama2.

%% file: latex/figtab_text/tab_ablation.tex
\begin{table}[t]
    \centering
    \small
    \tabcolsep 8pt
    \caption{Ablation Studies comparing different input modalities, 3D representation, pertaining strategy, and data augmentation on ScanQA and SQA3D benchmarks. \#Param reports the number of parameter of different feature extractors.}
    \label{tab:ablation}
    \begin{tabular}{ccccc | ccccccc}
    \toprule
    & \textit{Mod.} & \textit{Rep.} & \textit{Pre.} &  \textit{Aug.} & \#Param & \textbf{ScanQA}  & \textbf{SQA3D}  \\
    \midrule
     (a)&3D & obb  &  x &  x & -- &25.3 &    53.1  \\
     (b)&3D & tok-pc  &  x &  x & 24M & 24.9  &    52.7 \\
     (c)&3D & ds-pc  &  x &  x & -- &25.0  & 52.9   \\
     (d)&3D & Q-pc  &  x &  x &  105M &24.9  &    52.6\\
     (e)&3D & voxel  &  x &  x &  1.4M &25.0  &      52.8 \\ 
     (f)&3D & voxel  &  \checkmark &  x &  1.4M &25.6  &    53.2  \\ 
     (g)&3D & voxel  &  \checkmark &  \checkmark &  1.4M &25.6   &  53.6\\ 
     \bottomrule
    \end{tabular}
\end{table}

%% file: latex/figtab_text/tab_ablation_plan.tex
\begin{table}[t]
    \centering
    \small
    \tabcolsep 12pt
    \renewcommand\arraystretch{1.}
    \captionsetup{width=1.0\textwidth} 
    \caption{High-level planning accuracy(HLP) on Alfred dataset valid unseen/seen set with different inference strategy. Full model outperform strategies without egocentric and scene state updates.
    }
    \label{tab:ablation_plan}
    \begin{tabular}{lcc}
    \toprule
    {Model} & 
    {Valid Unseen} & {Valid Seen}  \\
    \midrule
    w/o egocentric & {72.64}&{81.20} \\
    w/o scene & {77.22}& {82.11} \\
    \methodname & {79.63}& {82.78} \\
     \bottomrule
    \end{tabular}
    \vspace{-0.2cm}
\end{table}

%% file: latex/figtab_text/fig_4_ds.tex
\begin{figure}[t]
    \centering
    \begin{subfigure}[b]{0.36\textwidth}
        \includegraphics[width=\textwidth]{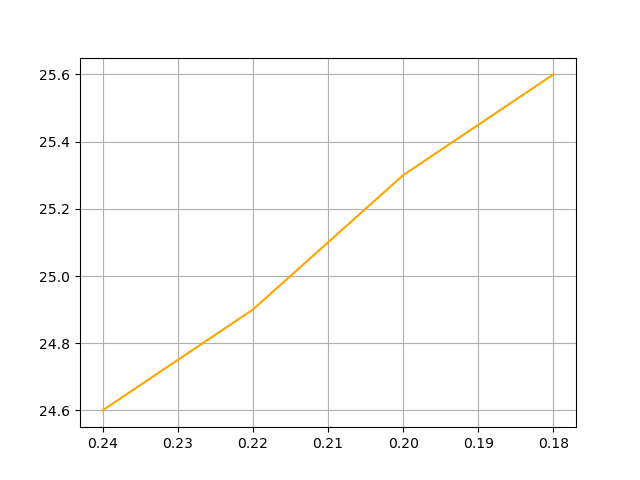} 
        \caption{ScanQA-Resolution}
        \label{fig:scanqa}
    \end{subfigure}
    \begin{subfigure}[b]{0.36\textwidth}
        \includegraphics[width=\textwidth]{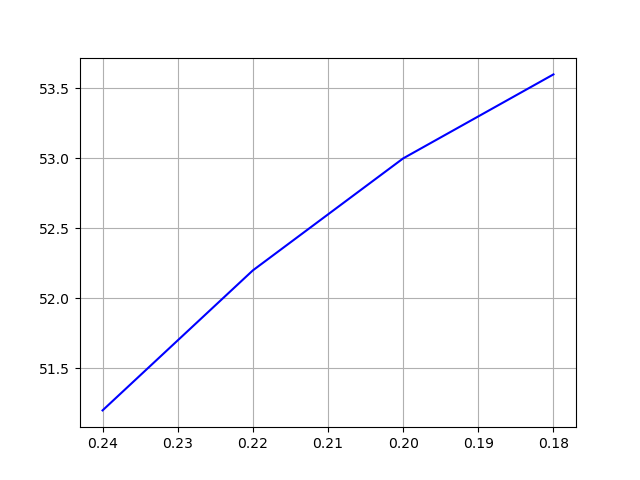} 
        \caption{SQA3D-Resolution}
        \label{fig:sqa3d}
    \end{subfigure}
    \begin{subfigure}[b]{0.26\textwidth}
        \includegraphics[width=\textwidth]{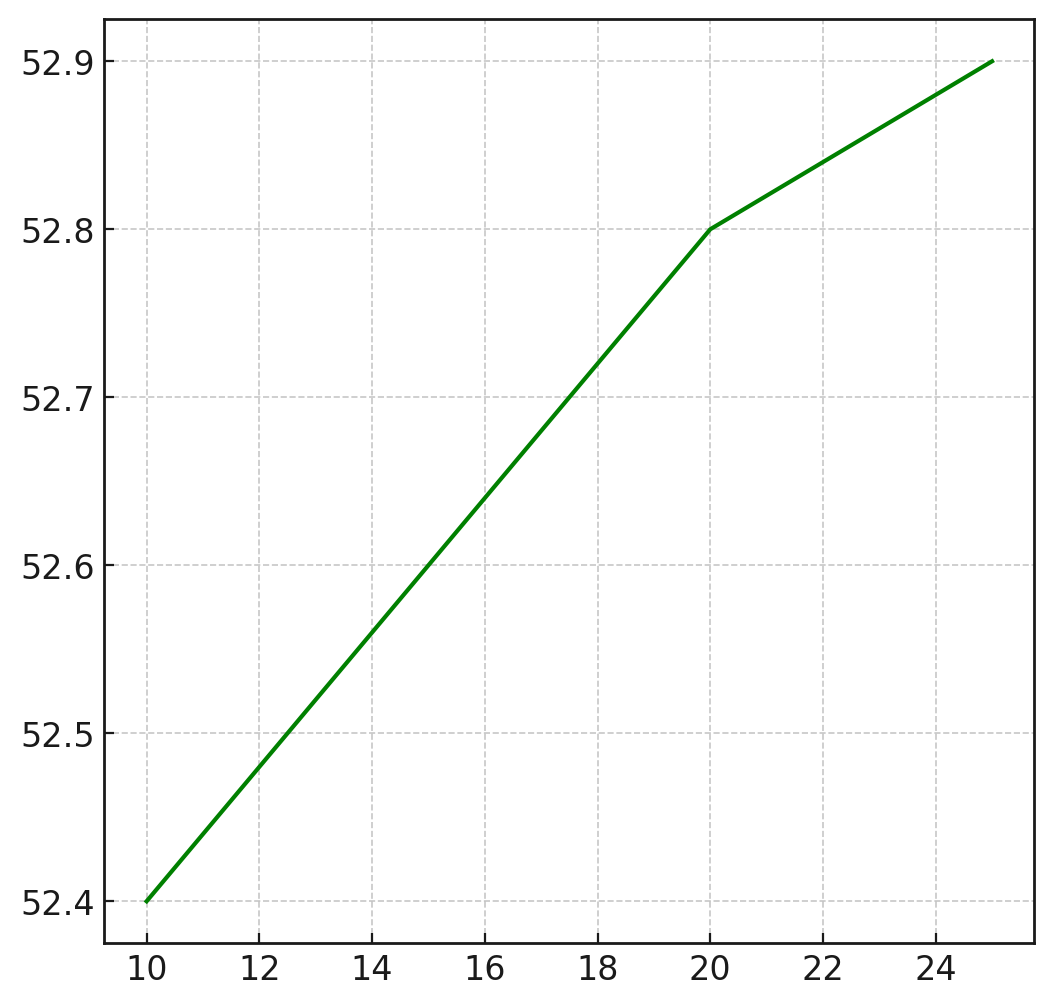} 
        \caption{SQA3D-\#views}
        \label{fig:sqa3d}
    \end{subfigure}
    \caption{EM scores on the Y-axis with decreasing voxel grid intervals (X-axis), indicating higher resolution leads to better performance. EM score slightly increases with number of views increases.}
    \label{fig:ds}
\end{figure}

%% file: latex/sections/6_conclusion.tex
\section{Conclusion}
We introduces \methodname{}, a 3D-visual-language model designed for understanding and reasoning in interactive 3D scenes. The model enhances LLMs with both egocentric and geo-centric 3D spatial understanding, leading to top-tier performance in 3D VQA and interactive planning. Our hybrid feature representation effectively captures comprehensive spatial information and facilitates dynamic state updates. We also present our 3D-visual-language dataset, highlighting the beneficial use of frame data in concept alignment. However, \methodname{} faces limitations such as LLM input token length, challenges in processing dynamic scenes without a state detector, lacking geometry feature, and language hallucinations. Despite these challenges, \methodname{} represents an advancement in 3D visual understanding, paving the way for more complex agent interactions in indoor settings.

%% file: latex/supp/all.tex
\begin{center}
    \Large Supplementary for Scene-LLM
\end{center}
\input{latex/supp/text/00_intro}
\input{latex/supp/text/3_more_results}
\input{latex/supp/text/0_prompt}
\input{latex/supp/text/4_training}

%% file: latex/supp/text/00_intro.tex
This document serves as the supplementary material for \textit{\methodname: Extending Language Models for 3D Visual Understanding and Reasoning}. Due to space constraints in the main paper, this supplementary section provides additional experimental results. It includes extended examples of interactive planning, detailed benchmark results for dense caption generation, and further ablation studies that explore various modalities and the impact of using frame data. Additionally, we present comprehensive information on the generation of frame data and scene data, encompassing prompts, post-processing steps, and data distribution specifics. Finally, detailed explanations of the training and inference methodologies as outlined in the main paper are also included for thorough understanding.

\newpage

%% file: latex/supp/text/3_more_results.tex

\section{Ablation Studies comparing different modalities.}
\input{latex/supp/figtab_text/tab_ablation_modality}
Our study evaluates performance on the ScanQA\cite{azuma2022scanqa} and SQA3D\cite{ma2022sqa3d} benchmarks across various modalities, including text, video, Bird's Eye View (BEV), and 3D. In the text modality, we employ the Llama-2-7b\cite{touvron2023llama} backbone, while video and BEV modalities share identical network architectures and training strategies as \methodname{}. Within the text modality, we compared object bounding boxes and object entities, finding the latter more effective.

We conducted experiments (a-c) using the text modality, encompassing zero-shot inference, fine-tuning with scene data, and additional task-specific tuning. Experiments (d-f) explored video, BEV, and 3D modalities. Our findings reveal that video and BEV modalities outperform the text modality in the ScanQA benchmark but underperform in SQA3D. Both methods underperform 3D data on both benchmarks. This suggests that spatial downsampling methods may not be ideal for processing video and image patches, underscoring the value of 3D information in preserving spatial knowledge more effectively.

\section{Ablation Studies comparing concept numbers.}
\input{latex/supp/figtab_text/fig_concept_number}
Fig.~\ref{fig:concept_count} presents a detailed analysis comparing the richness of vocabulary and conceptual variety between scene data and frame data, as well as their impact on model training. The analysis focuses on two key metrics: the number of unique words and the number of unique nouns found in the captions of both datasets, excluding stopwords. The scene data, consisting of 200,000 captions, is compared with frame data, which contains 190,000 captions.

In terms of unique words, the metric refers to the total count of distinct words across the dataset, providing insight into the lexical diversity. Similarly, the count of unique nouns, excluding stopwords, gives an indication of the variety and specificity of concepts covered in the data. The figure reveals that frame data exhibits a higher count of both unique words and nouns, suggesting a more diverse and conceptually rich dataset.

Furthermore, the figure also compares the output when the model is trained on these two datasets, using 142 validation scenes from the ScanNet V2 dataset\cite{Dai_2017_CVPR}. Notably, the model trained with frame data generates a greater number of nouns during inference, indicating that frame data effectively imparts a broader range of fine-grained concepts to the model. This enhanced conceptual diversity likely contributes to more robust and nuanced model performance in tasks requiring detailed understanding and reasoning.

\section{Ablation Studies comparing convergence speed.}
\input{latex/supp/figtab_text/fig_convergence}
Fig.~\ref{fig:convergence} displays a comparative analysis of the training loss trajectories during the first 6,000 steps of the pretraining phase, as specified in \cite{touvron2023llama}. This phase emphasizes concept alignment, leveraging two distinct datasets: frame data, which is accompanied by 190,000 textual annotations, and scene data, with 200,000 textual annotations.

The graph clearly shows a quicker reduction in training loss for the model utilizing frame data as opposed to scene data. This quicker convergence when using frame data suggests that the frame dataset provides a richer and more diverse set of concepts for the model to learn. This diversity likely facilitates more efficient learning and understanding, enabling the model to more rapidly adjust its parameters for optimal performance. Additionally, the swifter convergence with frame data could be linked to the nature of the 3D data itself. Frame point set typically contains more detailed and precise information that aligns closely with its accompanying annotations. In contrast, the scene point set, being a downsampled representation, may not align as closely with its textual annotations. This misalignment could lead to slower learning as the model struggles to correlate the visual data with its textual descriptions, thereby explaining the observed difference in convergence rates.

\section{Result on Dense Caption Generation.}
\input{latex/supp/figtab_text/tab_dense_caption}
We detail our investigation into dense caption generation on the Scan2Cap benchmark\cite{chen2021scan2cap}, using task-specific fine-tuning. This benchmark evaluates the ability of models to generate detailed captions for objects within a 3D scene, based on their bounding box inputs. Similar to the 3D-LLM\cite{hong20233dllm}, the model receives the center and dimensions of an object's bounding box as input, formatted as \texttt{[position x, position y, position z, length x, length y, length z]}, with each dimension represented as a two-digit floating-point number.

In our study, \methodname{} is compared against both a modular method outlined in \cite{chen2021scan2cap} and another 3D-visual-language model from \cite{hong20233dllm}. The results demonstrate that \methodname{} outperforms the other methods across all evaluated metrics, indicating \methodname{}'s proficiency in comprehending 3D scenes and accurately understanding spatial coordinates when fine-tuned for specific tasks. The result indicates that when \methodname{} is combined with 3D grounding methods, it holds the potential to tackle a wide array of complex tasks. 

\section{Result on 3DMV-VQA.}
\input{latex/supp/figtab_text/tab_3dmv}
We evaluated the performance of \methodname{} on the 3DMV-VQA benchmark\cite{hong20233dllm}, a benchmark specifically designed for testing a model's ability to understand and reason about 3D scenes, using data from the Habitat Matterport 3D Dataset\cite{ramakrishnan2021habitat}. 

For our evaluation, we selected a subset of the 3DMV-VQA benchmark that aligns with our model's focus on single-room scenarios. This subset, sourced from the benchmark's open-source codebase, comprises a total of $1,212$ scenes. Notably, the answers in this benchmark are generally concise, typically consisting of one word or a short phrase. To align with this format, we employed the identifier \texttt{"Answer the question using one word or one phrase"} in our zero-shot (zs) evaluation setup.

We assessed \methodname{} under two distinct settings: zero-shot and task-specific tuning. In the zero-shot setting, where no additional fine-tuning was applied, \methodname{} demonstrated reasonable performance, indicating an inherent capability to understand and respond to queries related to 3D scenes. However, when further optimized through task-specific tuning, the model's performance showed a notable improvement.

%% file: latex/supp/figtab_text/tab_ablation_modality.tex
\begin{table}[th]
    \centering
    \small
    \tabcolsep 8pt
    \caption{Ablation Studies comparing different input modalities. Including textual zero-shot, textual fine-tuned, and textual tasks-specific-tuned, video, bird-eye views, and 3D point sets. }
    \label{tab:ablation}
    \begin{tabular}{ccc | cccccc}
    \toprule
    & \textit{Mod.} & \textit{Tune.} & \textbf{ScanQA}  & \textbf{SQA3D}  \\
    \midrule
     (a)&text & Zero-Shot  & 13.1  &    34.1   \\
     (b)&text & Finetune   & 17.9  &    48.9   \\
     (c)&text & Task Tune   & 18.0  &    49.5   \\
     (d)&Video & Task Tune  & 18.9  &    48.2   \\
     (e)&BEV & Task Tune   & 19.2  &      48.4   \\
     (f)&3D & Finetune  & 25.0  &      52.8 \\ 
     \bottomrule
    \end{tabular}
\end{table}

%% file: latex/supp/figtab_text/fig_concept_number.tex
\begin{figure}[h]
    \centering
    \includegraphics[width=0.49\textwidth]{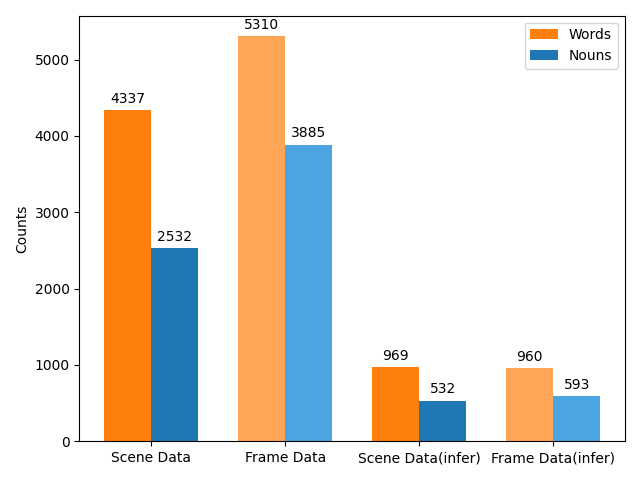} %
    \caption{Comparing the number of unique words and nouns in the scene dataset and frame dataset. Comparing the number of unique words and nouns inferred by training with the model with scene data and frame data.}
    \label{fig:concept_count}
\end{figure}

%% file: latex/supp/figtab_text/fig_convergence.tex
\begin{figure}[h]
    \centering
    \includegraphics[width=0.49\textwidth]{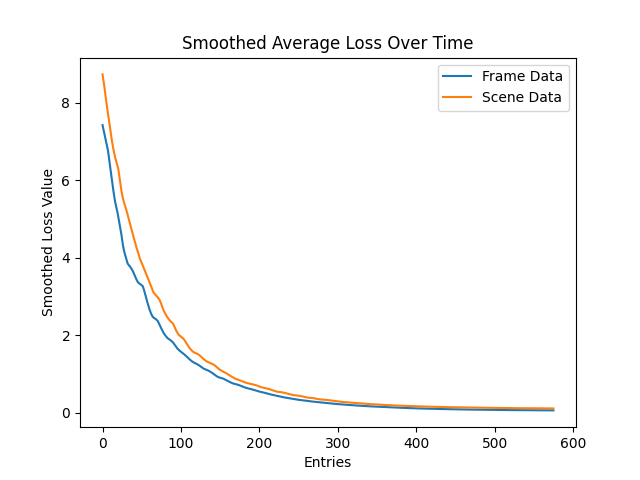} %
    \caption{Loss curves for training the projection layer at the first $6k$ iterations. Using frame data converges faster than using scene data.}
    \label{fig:convergence}
\end{figure}

%% file: latex/supp/figtab_text/tab_dense_caption.tex
\begin{table}[h]
    \centering
    \small
    \tabcolsep 8pt
    \caption{Result on 3D Dense Caption Benchmark Scan2Cap\cite{chen2021scan2cap}. }
    \label{tab:caption}
    \begin{tabular}{ccccc}
    \toprule
    & \textbf{CIDEr} & \textbf{BLEU-4} & \textbf{METEOR}  & \textbf{ROUGE}  \\
    \midrule
     Scan2Cap\cite{chen2021scan2cap} & 35.2 & 22.4 & 21.4 & 43.5   \\
     3D-LLM\cite{hong20233dllm} & -- & 8.1 & 13.1 & 33.2   \\
     \methodname{} & 37.9 & 24.1   & 21.8  & 45.6   \\
     \bottomrule
    \end{tabular}
\end{table}

%% file: latex/supp/figtab_text/tab_3dmv.tex
\begin{table}[h]
    \centering
    \small
    \tabcolsep 8pt
    \caption{Result on 3DMV-VQA(single room only) Benchmark. }
    \label{tab:caption}
    \begin{tabular}{cccccc}
    \toprule
    & \textbf{Concept} & \textbf{Counting} & \textbf{Relation} & \textbf{Comparison} & \textbf{Overall}  \\
    \midrule
     \methodname{}(zs) &  69.9 & 31.2 & 62.0 & 73.5 & 59.6   \\
     \methodname{}(tt) & 70.2 & 33.5 & 62.4 & 75.1 & 61.3  \\
     \bottomrule
    \end{tabular}
\end{table}

%% file: latex/supp/text/0_prompt.tex
\section{3D Dataset}
Our 3D scene dataset is composed of three parts: we collect $564$ training scenes from ScanNet-V2\cite{Dai_2017_CVPR}, $1212$ scenes from HM3D\cite{ramakrishnan2021habitat} dataset, and $7075$ randomly configued scenes from iThor\cite{Kolve2017AI2THORAn} dataset. There are in total of $8851$ scenes. For the 3D frame, we use Habitat simulator and Ai2Thor simulator to collect frame data. We also use the multi-view images from ENet\cite{paszke2016enet} for ScanNet frame data. For each scene, we collect $20$ frames from multiple views. 

\section{Frame Data Generation Details.}
\subsection{Prompts}
We use MiniGPT-V2\cite{chen2023minigpt} for frame data generation. We use the identifier \texttt{[grounding]} as we found it can effectively reduce hallucination in the generated texts. Here are the prompts we use: 
\begin{enumerate}
    \item \texttt{[grounding] Describe this image in detail. Give as many details as possible. Say everything you see.}
    \item \texttt{[grounding] Describe the objects in this image. }
\end{enumerate}

The first prompt generates dense captions, and the second prompt generates object-centric descriptions. We filtered out outputs with meaningless characters, and a total of around $190k$ textual descriptions are collected.

\section{Scene Data Generation Details.}
\subsection{Prompts}
In the section, we provide the prompts used for scene data generation. The prompts we provided are composed of two parts $P_{context}$ and $P_{instruction}$, where $P_{context}$ introduces the scene context used for data generation and $P_{instruction}$ indicates which type of instructional data to generate. 

For the scene context, we include four types of context. Here are the context types and the corresponding $P_{context}$:
\begin{enumerate}
    \item \textbf{Dense Caption:} \texttt{Given a script starting with "Scripts:", where a person describes objects in an indoor scene ... }
    \item \textbf{Object Caption: } \texttt{Based on a script beginning with "Scripts:" detailing objects in an indoor scene ...}
    \item \textbf{Object Entites: } \texttt{Using the "OBJECT LIST:", which describes objects and their bounding boxes in an indoor environment in the format: object [position x, position y, position z, length x, length y, length z] ...}
    \item \textbf{Object bounding boxes: } \texttt{Provided is a list titled "OBJECT LIST:", which details objects and their bounding boxes in an indoor setting, formatted as: object [position x, position y, position z, length x, length y, length z] ... }
\end{enumerate}

We generate $12$ types of instruction following annotations. In order to reduce hallucination, we only use some types of contexts for each type of data generation. Here are the annotation types, the corresponding $P_{instruction}$ and the used context types.:
\begin{enumerate}
    \item \textbf{Scene Caption:} \texttt{provide a summarization starting with "Summary:". The summary should detail the objects, their positions, appearances, and the function of the room. Scripts:} Used contexts: scene caption, object entities, object bounding boxes. 
    \item \textbf{Object Caption:} \texttt{provide a summarization starting with "Summary:". The summary should detail the objects, their positions, appearances, and the function of the room. Scripts:} Used contexts: object caption.
    \item \textbf{General Question Answering: } \texttt{visualize a scenario with a Human and a robot assistant present. Construct a Question-Answer pair, where the Human asks and the robot answers. Format as follows: Q: [Human's question based on the script details] A: [Robot's answer based on the script]}. Used contexts: scene caption, object caption, object entities, object bounding boxes. 
    \item \textbf{Question Answering(Concept): } \texttt{Produce question-answer pairs wherein the human inquires about the existence of the objects. Q: [Human's question about the existence of objects] A: [Robot's response based on the script details].} Used contexts: scene caption, object entities, object bounding boxes. 
    \item \textbf{Question Answering(Concept-Negation): } \texttt{Generate question-answer pairs where the human asks about objects that are not present in the given list. The answers should confirm the absence of these objects. Q: [Human's question about the existence of objects] A: [Robot's response based on the script details].} Used contexts: scene caption, object entities, object bounding boxes. 
    \item \textbf{Question Answering(Counting): } \texttt{Produce question-answer pairs wherein the human inquires about the number of the objects. Q: [Human's question about the number of objects] A: [Robot's response based on the script details].} Used contexts: object entities, object bounding boxes. 
    \item \textbf{Question Answering(Spatial): } \texttt{picture a setting with a Human and a robot assistant. Create question-answer pairs wherein the Human inquires about the location of specific objects. Frame your dialogue in this manner: Q: [Human's question about an object's position. A: [Robot's answer referring to the object list].} Used contexts: scene caption, object bounding boxes. 
    \item \textbf{Question Answering(Comparison): } \texttt{Produce question-answer pairs wherein the human inquires about differentiating aspects of the objects. The queries should compare the number of the objects./The queries should compare the size of the objects./The queries should compare the functionalities of the objects./The queries should emphasize the distinctions between the objects.  Please structure the dialogue as follows: Q: [Human's question distinguishing the objects from the script] A: [Robot's response based on the script details].} Used contexts: scene caption, object captions, object entities, object bounding boxes. 
    \item \textbf{Question Answering(Navigation): } \texttt{envisage a scenario where a robot navigates the area based on human instructions. Generate question-answer pairs that relate to navigating within the scene. The robot may inquire about directions, routes, and objects encountered along the way. Please adhere to the following format: Q: [Robot's question about navigation relative to the objects in the script] A: [Human's answer guiding the robot, using information from the script].} Used contexts:  scene caption, object bounding boxes. 
    \item \textbf{Human-robot Dialogue: } \texttt{envision a scenario with a Human and a robot assistant interacting. Create a multi-round dialogue between them. Begin the human's lines with "Human:" and the robot's lines with "Robot:"."} Used contexts: scene caption, object entities, object bounding boxes. 
    \item \textbf{Task Decomposition: } \texttt{suggest daily tasks or chores relevant to the described environment. Each task should include a concise description and a step-by-step guide on how to complete it using only objects mentioned in the scene. Answer using this format: Task: [Brief description of the task Step-by-step instruction: 1.[First step] 2.[Second step] 3.[Third step].} Used contexts: scene caption, object entities, object bounding boxes. 
    \item \textbf{Functionality Improvement: } \texttt{ suggest ways to enhance the functionality of the environment. Your recommendation should focus on improvements like better lighting for reading. Provide your answer in the given format: Function: [Aspect of the environment to enhance] Method: [Suggestion for improvement based on the described scene].} Used contexts: scene caption, object entities, object bounding boxes. 
\end{enumerate}

\subsection{Data Post-processing}

To process the instructional data generated from stage 2, we use LLama-2-chat-70b\cite{touvron2023llama} to remove any redundant and inaccurate information. Here is the prompt we use at this stage: \texttt{Correct errors in the provided paragraph. Eliminate repeated sentences, meaningless characters, and non-English phrases. Remove unnecessary repetition. Complete incomplete sentences. Provide the corrected answer directly without additional explanation.} 

\subsection{Instruction-type Distribution}
\input{latex/supp/figtab_text/fig_data_distribution}
Fig.~\ref{fig:distribution} shows the distribution of each type of instructional-following annotations, including question answering(QA), object caption, tasks decomposition, dense caption, human-robot dialogue and function improvement. 

%% file: latex/supp/figtab_text/fig_data_distribution.tex
\begin{figure}[h]
    \centering
    \includegraphics[width=0.49\textwidth]{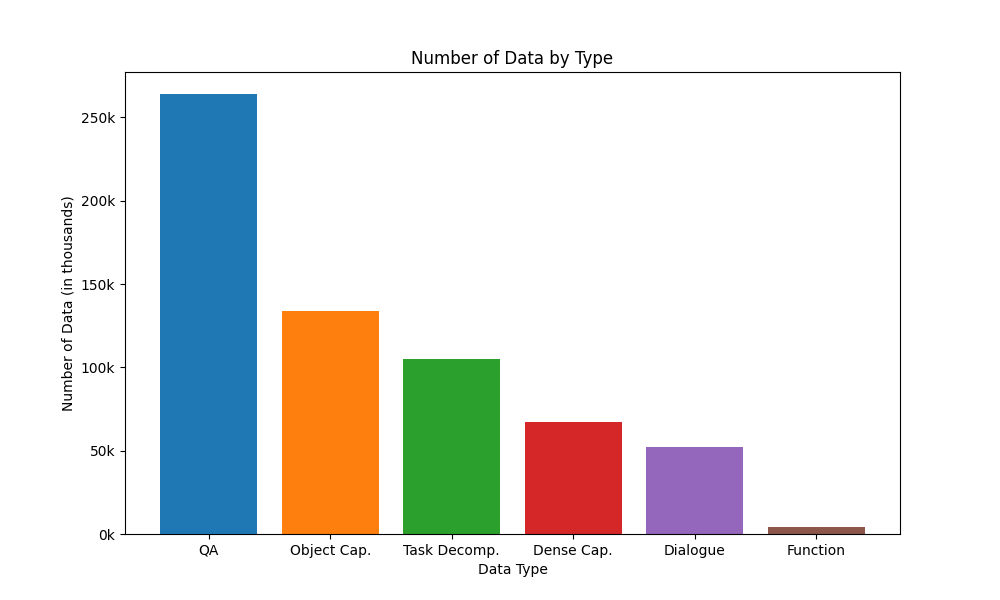} %
    \caption{Distribution of the scene data by each type of instruction.}
    \label{fig:distribution}
\end{figure}

%% file: latex/supp/text/4_training.tex
\section{Training Detail}
\subsection{Projection Layer Structure.}
The feature of each point is a concatenation of CLIP-H/14 feature, point color, and point position, which is of $1030$ dimension. The input dimension to the LLM is $768$ dimension. The projection layer is composed of a fully connected layer of a matrix shape $[1030, 768]$, followed by a GELU activation layer, and then is a fully connected layer of a matrix shape $[768, 768]$.

\subsection{Pre-training Projection Layer.}
We use $32 \times$ NVIDIA A100 GPU to train the projection layer. The total batch size is $64$ at this stage. We use AdamW as the optimizer, and a learning rate of $1e-5$ to train the projection layer. The training starts with $1000$ steps of warmup with a warmup learning rate of $1e-6$. The training takes a total of $5$ hours. 

\subsection{Finetuning the Projection Layer with the LLM.}
We use $32 \times$ NVIDIA A100 GPU to finetune the projection layer with the LLM. The total batch size is $64$ at this stage. We use AdamW as the optimizer, and a learning rate of $2e-5$ to finetune the model. The training starts with $2000$ steps of warmup with a warmup learning rate of $1e-6$. The fine-tuning takes a total of $8$ hours. 

\subsection{Task-specific Finetuning.}
We use the same setting for all benchmarks for task-specific tuning. We use $32 \times$ NVIDIA A100 GPU to finetune the projection layer with the LLM. The total batch size is $64$ at this stage. We use AdamW as the optimizer, and a learning rate of $2e-5$. For the tuning steps, we report the result of $1,500$ step for ScanQA and SQA3D, $3,500$ step for Alfred, $6,000$ step for Scan2Cap, and $5,000$ step for 3DMV-VQA.